\documentclass[conference]{IEEEtran}
\IEEEoverridecommandlockouts
\usepackage{cite}
\usepackage{amsmath,amssymb,amsfonts}
\usepackage{algorithmic}
\usepackage{graphicx}
\usepackage{textcomp}
\usepackage{xcolor,booktabs}
\usepackage{url}
\usepackage{subcaption}

\usepackage{amsmath}

\usepackage{soul}
\usepackage{derivative}
\usepackage{hyperref}
\usepackage{easyReview}

\def\BibTeX{{\rm B\kern-.05em{\sc i\kern-.025em b}\kern-.08em
    T\kern-.1667em\lower.7ex\hbox{E}\kern-.125emX}}
\begin{document}

\title{H-Alpha Anomalyzer: An Explainable Anomaly Detector for Solar H-Alpha Observations}

\author{
    \IEEEauthorblockN{Mahsa~Khazaei\IEEEauthorrefmark{1},
    Azim~Ahmadzadeh\IEEEauthorrefmark{1},
    Alexei~Pevtsov\IEEEauthorrefmark{2}
    Luca~Bertello\IEEEauthorrefmark{2}
    Alexander~Pevtsov\IEEEauthorrefmark{2}}
    \IEEEauthorblockA{
        \IEEEauthorrefmark{1}Department of Computer Science,
        University of Missouri-St. Louis,
        \IEEEauthorrefmark{2}National Solar Observatory, 
        Email: \IEEEauthorrefmark{1}mkdgy@umsl.edu \\
        \vspace{-1.1cm}
    }
}

\maketitle

\begin{abstract}
The plethora of space-borne and ground-based observatories has provided astrophysicists with an unprecedented volume of data, which can only be processed at scale using advanced computing algorithms. Consequently, ensuring the quality of data fed into machine learning (ML) models is critical. The H$\alpha$ observations from the GONG network represent one such data stream, producing several observations per minute, 24/7, since 2010. In this study, we introduce a lightweight (non-ML) anomaly-detection algorithm, called H-Alpha Anomalyzer, designed to identify anomalous observations based on user-defined criteria. Unlike many black-box algorithms, our approach highlights exactly which regions triggered the anomaly flag and quantifies the corresponding anomaly likelihood. For our comparative analysis, we also created and released a dataset of 2,000 observations, equally divided between anomalous and non-anomalous cases. Our results demonstrate that the proposed model not only outperforms existing methods but also provides explainability, enabling qualitative evaluation by domain experts.

\end{abstract}


\section{Introduction}
Millions of H$\alpha$ images are produced by the NSF's Global Oscillation Network Group (GONG, \cite{hill2018global}), some of which show anomaly patterns and can't be used by either algorithms or scientists. Detecting these images and removing them from the pipeline is a laborious task, especially on such a large scale. 
Motivated by this, we aim to analyze different solutions for cleaning corrupt H$\alpha$ images and do a comprehensive quantitative and qualitative analysis of their performance. But before doing so, we have to define what images are corrupt.

When anomalies are discussed, one of three scenarios is implied: 1) extreme cases that show undesired patterns like defects, 2) cases that are out-of-distribution which are sought after because they often correspond to rare or previously unseen phenomena that may carry valuable insights or indicate unusual conditions, 3) extreme cases that are unknown and we don't have information about \cite{ruff_2021_unifying}. The common thread between them is that they're much more scarce than normal data; hence, we don't have many examples available of them. The corrupt H$\alpha$ images discussed in this paper are of the first type; we know what type of anomalies we are looking for. The sources of these anomalies are discussed in Sec.~\ref{sec:haobs}.
Although solar events can be seen as events of interest and found using anomaly detection techniques, they are not the subject of this study.

\begin{figure}
    \centering
    \includegraphics[width=0.8
    \linewidth]{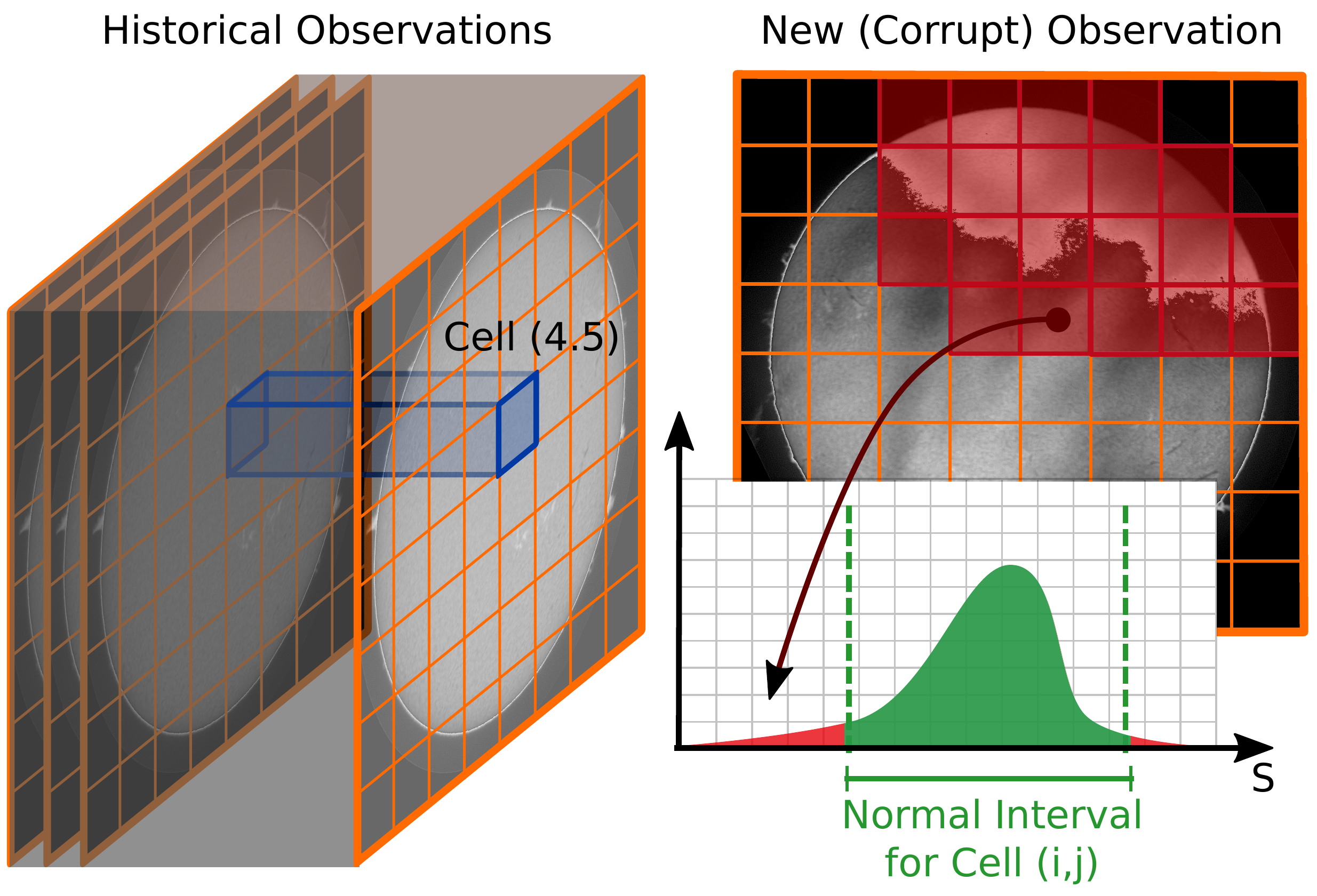}
    \caption{\footnotesize The figure shows the simple concept behind H-Alpha Anomalyzer. This anomaly-detection algorithm relies on the optimal range for average pixel values per grid cell. In this graphic, the distribution of the $S$ statistic, formed based on historical observations, is shown. The green section of the bell-shaped curve corresponds to normal (non-corrupt) cells of observations, whereas the red section corresponds to corrupt cells. The corrupt cells of the new observation are expected to fall outside the discovered ``acceptable'' range.}
    \label{fig:stacked_slices}
    \vspace{-0.4cm}
\end{figure}

In this work, we will introduce a novel statistical method for finding these images, H-Alpha Anomalyzer \cite{khazaei_h-alpha_2025}, which is available as a Python package. H-Alpha Anomalyzer is introduced for finding anomalous H$\alpha$ images. It works by dividing the image into patches on an $n\times n$ grid and finding the distribution of pixel intensities for each patch or ``cell" in the grid. As the H$\alpha$ images have high cadence, there is low variation of pixel intensity per cell in what is considered to be a normal image. Hence, the claim is that even a statistical approach would suffice to achieve a very effective anomaly detection algorithm. Fig.~\ref{fig:stacked_slices} shows historical observations and a corrupt observation side by side, and how the same cell has different pixel intensities in the anomalous image compared to the historical observations. Even the untrained eye can pick this difference between the pixel intensities in these two cells. We take advantage of the fixed position of the solar disk in GONG H$\alpha$ images to find the distribution of pixel intensities per cell for anomalous and non-anomalous images.

\section{GONG and H$\alpha$ Observations}
    Since our primary objective in this study is to present an algorithm for filtering corrupt H$\alpha$ observations captured by a network of ground-based observatories, we first introduce this network and the need for such data cleaning efforts.
    
    \subsection{Global Oscillation Network Group (GONG)}
        The GONG network consists of six identical telescopes strategically located around the globe to ensure near-continuous solar observations. The observatories are situated at Big Bear Solar Observatory (CA, USA), Mauna Loa Observatory (HI, USA), Learmonth Solar Observatory (Australia), Udaipur Solar Observatory (India), Observatorio del Roque de los Muchachos (Spain), and Cerro Tololo Inter-American Observatory (Chile). In addition, there are two engineering stations in Boulder, Colorado. This global distribution minimizes the gaps caused by the diurnal cycle and due to local weather. The telescopes are robotic: the observations start automatically based on the Solar ephemeries (computed coordinates of the Sun on the sky at the local site). The entrance optics is sealed to protect against light precipitations, and so, unless there is a major storm, the telescope may continue to operate even in partially cloudy conditions and occasional rain. Through this setup, GONG enables uninterrupted, year-round monitoring of the Sun \cite{hill1994global1, hill1994global2}. GONG was in operations since 1995, initially for studies in helioseismology and the magnetic field. In 2010, observations in the H$\alpha$ spectral line were added. The Time cadence of H$\alpha$ observations at each station is 60 seconds. In a study conducted in 2021, the network reported an average duty cycle of 93\%, representing the proportion of each 24-hour period during which data is collected. This high level of continuity has been sustained for the last 18 years of the 25-year period analyzed in \cite{jain2021continuous}.

    \subsection{H$\alpha$ Observations}
    \label{sec:haobs}
        An H$\alpha$ filter is specifically designed to isolate the narrow band (0.04 nm) of red light emitted by hydrogen atoms at a wavelength of 656.3 nanometers, a significant spectral line, mostly seen as absorption line on the solar disk, and as emission line in flares and off-limb prominences in the visible spectrum. The core of this spectral line is formed in the solar chromosphere, the layer of the solar atmosphere situated about 1,000-2,000 km above the visible solar surface (called photosphere). Images of the Sun showing dark sunspots correspond to the photosphere, while the chromospheric images show a different set of features, including dark elongated filaments and bright flares. Both erupting filaments and flares can be associated with the coronal mass ejections (CMEs) -- the main driver of space weather in the near-Earth environment. Thus, the observations in H$\alpha$ spectral line are important for monitoring the solar eruptive activity. To record rapid changes during the eruptive events, each GONG station takes observations at a 60-second cadence. The observation time is synchronized using the atomic clock signals from the GPS satellites, and the observing time at neighboring stations is offset by 20 seconds, thus, creating a network-merged time series of the chromospheric H$\alpha$ observations with 20-second cadence.
        
        The observations taken with H$\alpha$ filter allow ground-based observatories to capture highly contrasted images that reveal features not visible in white light, such as the intricate structures of solar filaments and the dynamic behavior of plasma guided by the Sun's magnetic field in the chromosphere. By focusing on this specific wavelength, H$\alpha$ observations enable scientists to monitor solar activity in near real-time, contributing valuable data for understanding solar dynamics, space weather prediction, and the broader impacts of solar variability on the Earth's environment.
        
        The observations are taken in an automatic regime, following a procedure described in detail by \cite{Diercke.etal2024}. During the observations, a solar disk tracker keeps the position of the Sun centered on the detector. For the final image, four separate frames taken over a short period of time are averaged together. This averaging of four frames could help with understanding the root cause of some image artifacts. For example, if the observations are taken with fast-moving clouds, one of the frames could catch an opening in the cloud, resulting in a bright spot in part of the image. Camera pixels in that part of the image could be oversaturated. For fast-moving objects (e.g., a commercial jetliner, birds, or even large insects), separate frames may catch their different positions (e.g., the plane's nose, the whole plane, and its tail). If the guider mistakenly moves the solar image during the exposure, the average of four frames would show several non-centered solar disks across the field of view.
        During the telescope maintenance, a technician may replace an actual image by an artificial one with an obvious pattern (e.g., a smiley face) as a warning sign that the data could be impacted by the work they are doing. The start and end time of the maintenance period is sent to the data center to exclude the artificial images, but mistakes are possible. When observations are taken with the Sun close to the horizon, man-made structures (e.g., buildings or relay towers) or tree branches could enter the field of view.

    \begin{figure}
        \begin{subfigure}{0.32\linewidth}
                \centering
                \includegraphics[width=1.0\linewidth]{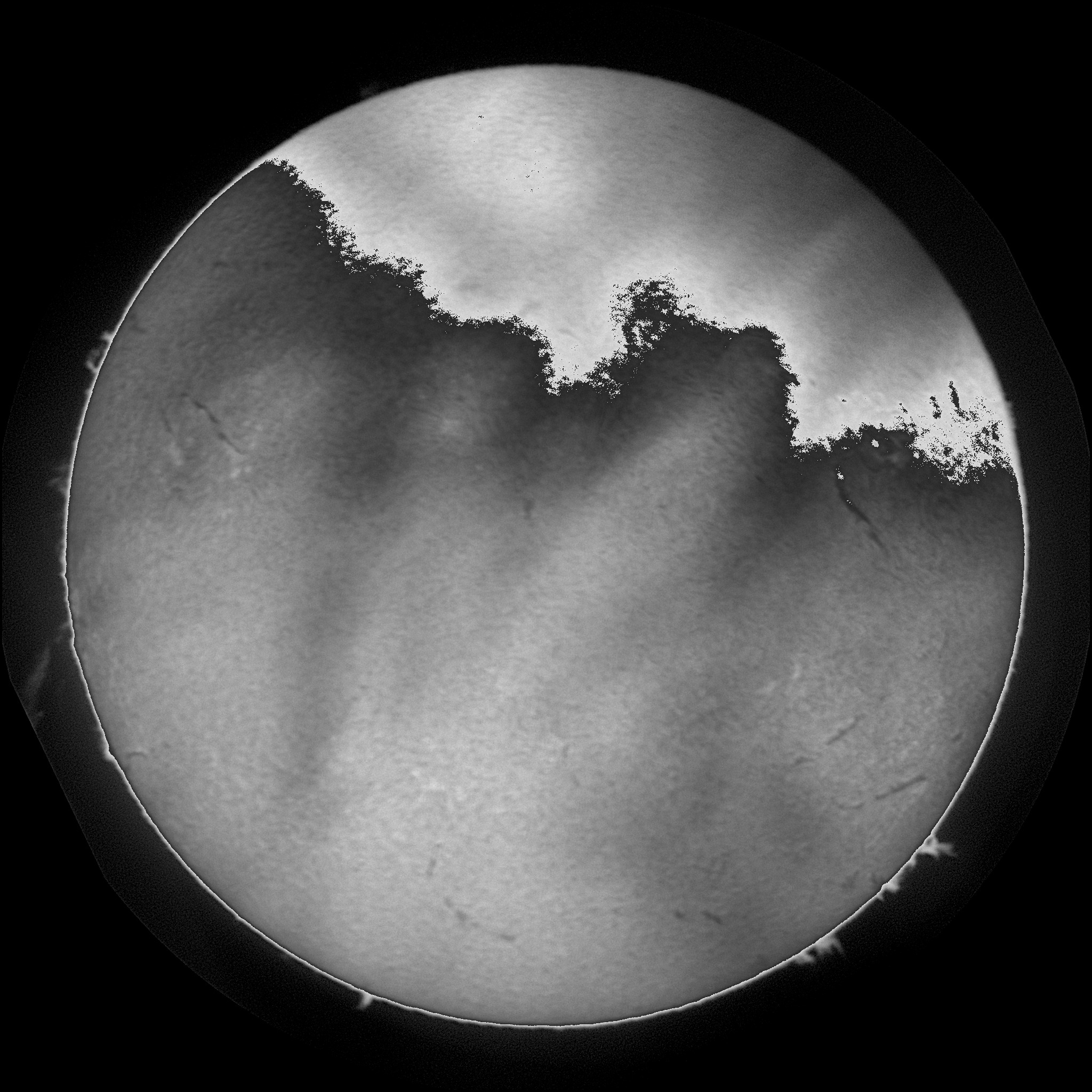}
                \caption{Over saturated}
                \label{fig:over_saturation}
        \end{subfigure}
        \begin{subfigure}{0.32\linewidth}
                \centering
                \includegraphics[width=1.0\linewidth]{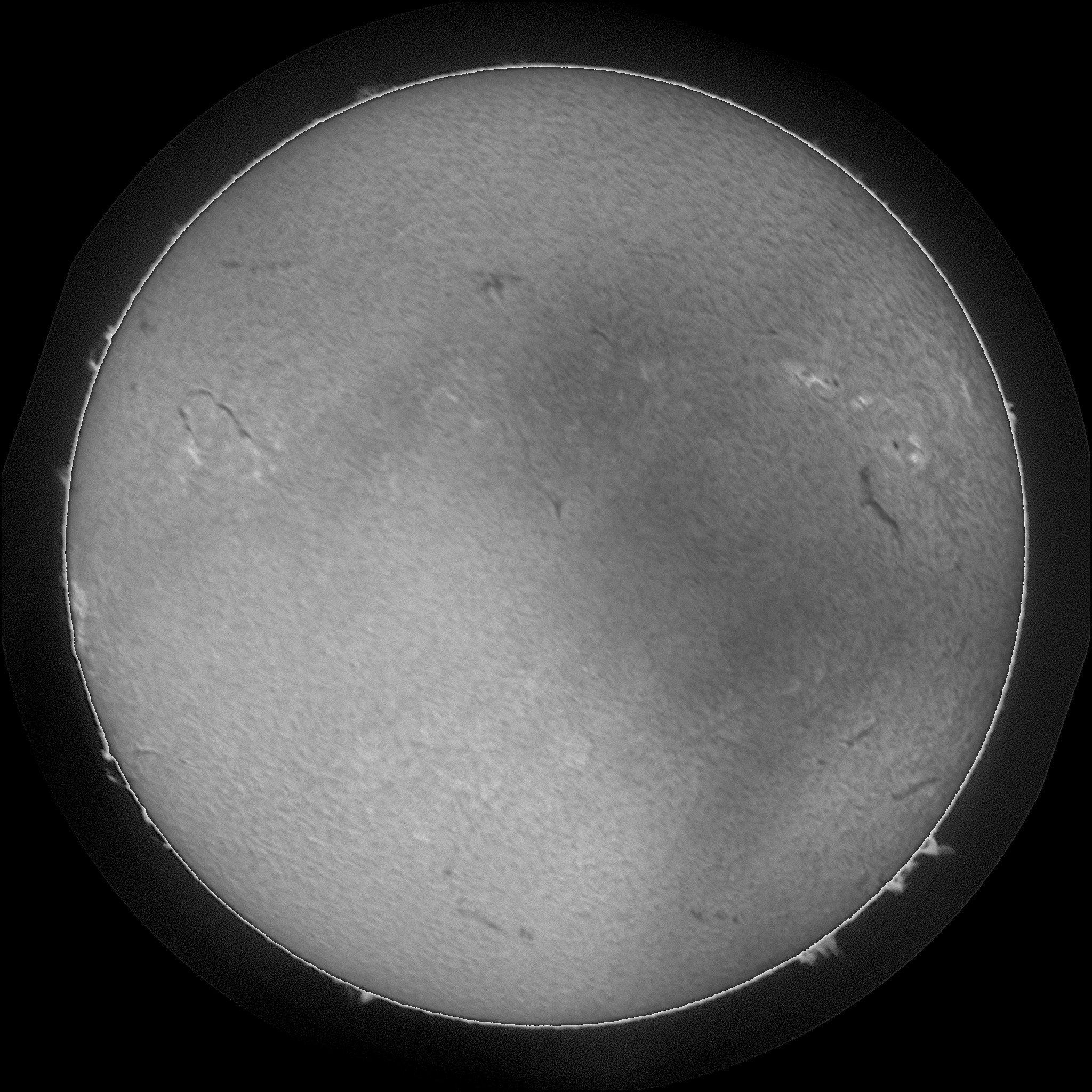}
                \caption{Cloudy}
                \label{fig:cloudy}
        \end{subfigure}
        \begin{subfigure}{0.32\linewidth}
                \centering
                \includegraphics[width=1.0\linewidth]{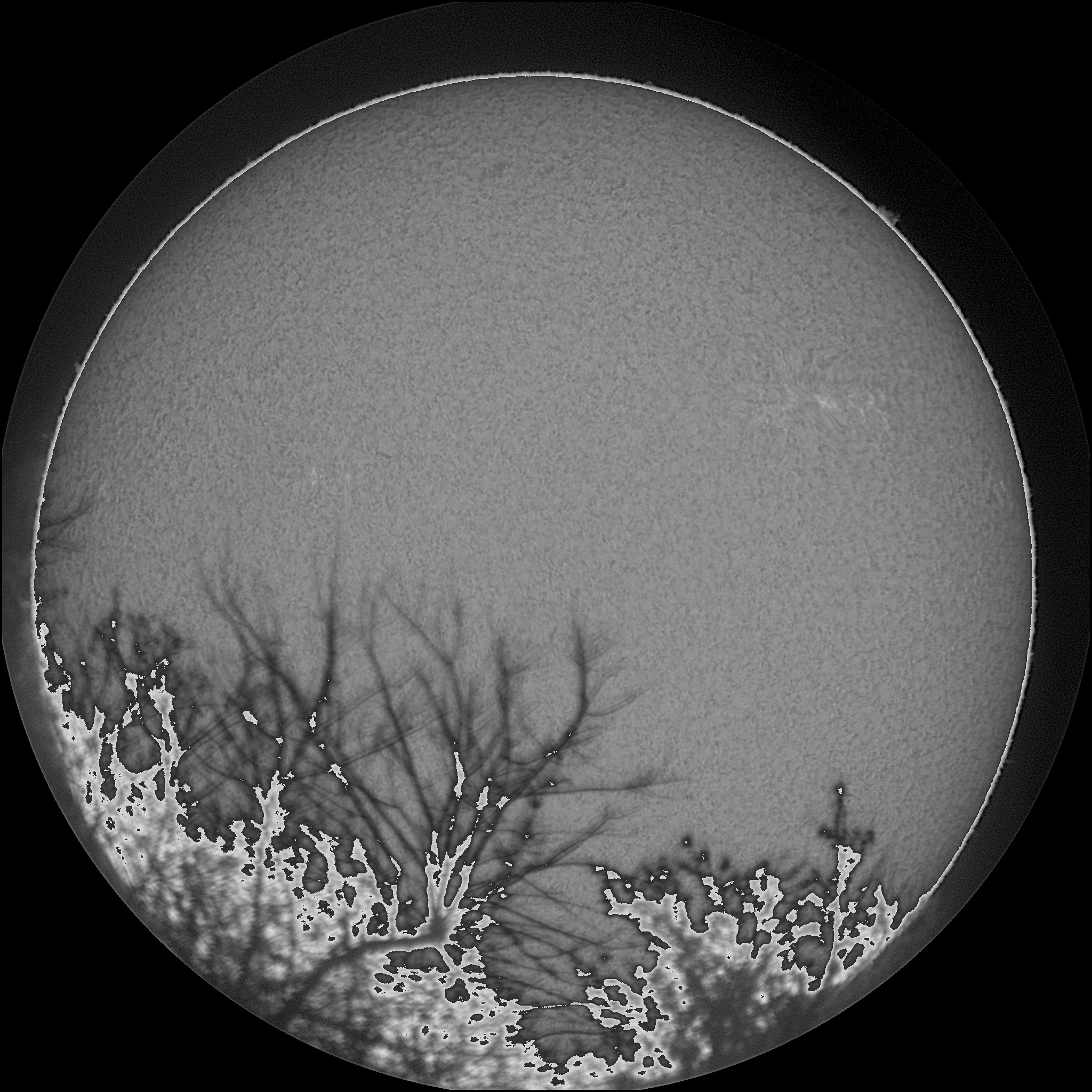}
                \caption{Obstructed}
                \label{fig:obstructed}
        \end{subfigure}

        \begin{subfigure}{0.32\linewidth}
                \centering
                \includegraphics[width=1.0\linewidth]{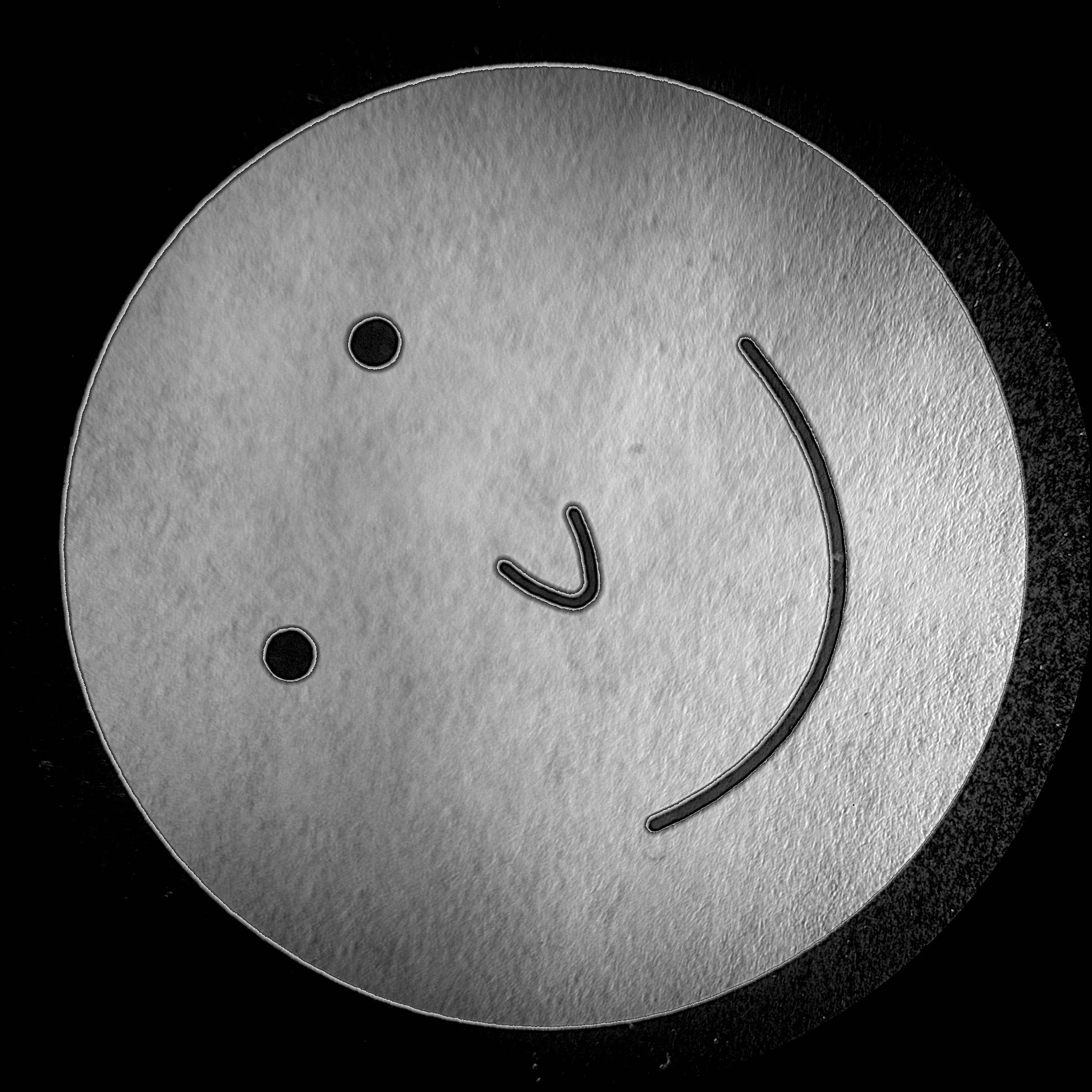}
                \caption{Filter on the lens}
                \label{fig:smiley}
        \end{subfigure}
        \begin{subfigure}{0.32\linewidth}
                \centering
                \includegraphics[width=1.0\linewidth]{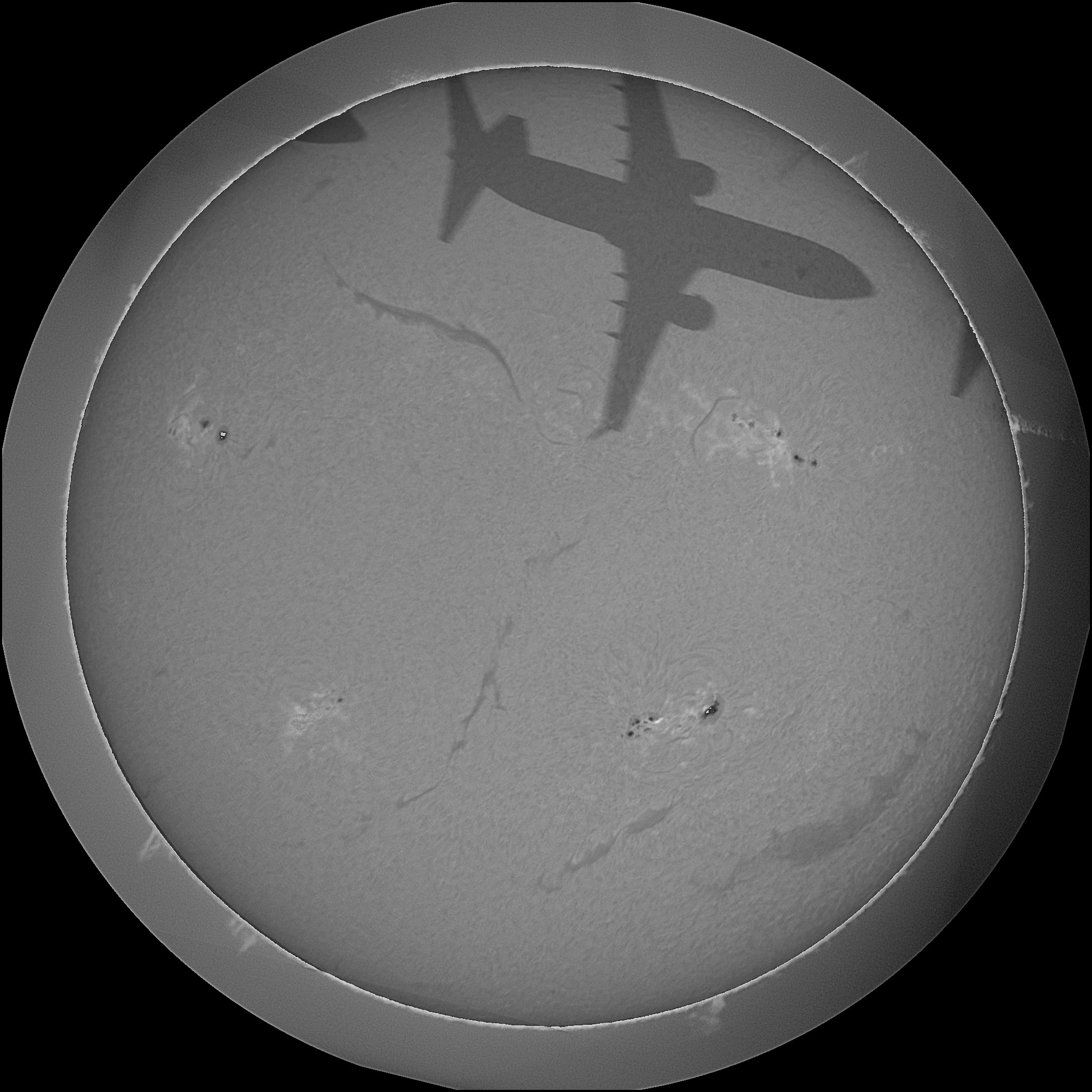}
                \caption{Obstructed}
                \label{fig:airplane}
        \end{subfigure}
        \begin{subfigure}{0.32\linewidth}
                \centering
                \includegraphics[width=1.0\linewidth]{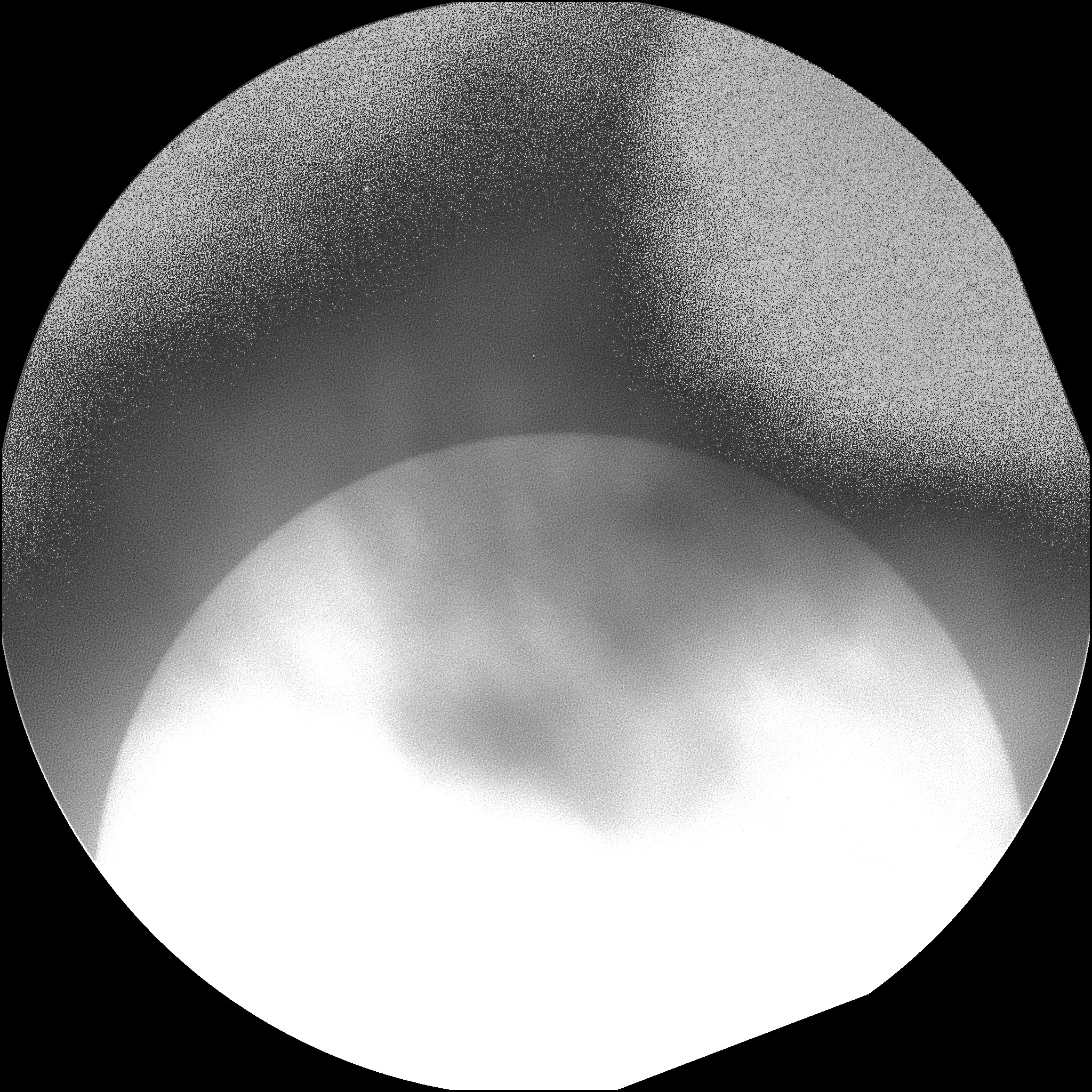}
                \caption{Uncentered disk}
                \label{fig:uncentered}
        \end{subfigure}
        
        \caption{\footnotesize A few examples of anomalous ground-based 
        observations. The observation timestamps from (a) to (f) 
        \mbox {20121214225734Lh, 20121215053634Lh, 20170306021250Uh,}
        \mbox {20130321222614Sh, 20220713164632Bh, 20180831000434Ah.}}
        \label{fig:anomaly_samples}
        \vspace{-0.5cm}
    \end{figure}

    \subsection{Lack of A Dedicated Quality Control}\label{subsec:lack_of_dedicated_quality_control}
        Despite the current popularity of H$\alpha$ data in scientific research, the GONG H$\alpha$ instrument was originally designed to provide high-cadence reference data rather than for scientific research \cite{hill2018global}. As a result, no dedicated data quality assurance pipeline was implemented to ensure the reliability of its products.
        The data quality control is implemented for GONG helioseismology and magnetogram observations, but not for  H$\alpha$ observations. Consequently, individual research groups have handled ``bad'' observations in an ad-hoc manner, leading to images with various anomalies. A few examples of such cases are shown in Fig.~\ref{fig:anomaly_samples}. Despite this, GONG’s success has made it a cornerstone for many scientific investigations. Since 2010, the National Solar Observatory (NSO) and the National Oceanic and Atmospheric Administration (NOAA) Space Weather Prediction Center (SWPC) have operationalized near-real-time processing of GONG data for monitoring solar activity and modeling solar wind and CMEs. Given its widespread use, the development of a robust and consistent quality-control pipeline is now critically needed. Given the large amount of observations (i.e., taken with 60-second time cadence at each of 6 GONG sites), the quality control needs to be automated.

\section{Data Collection}\label{sec:dataset}
    In order to take advantage of supervised algorithms, and also to evaluate the performance of the anomaly detection algorithms, we need a dataset of ``good" and ``bad" observations, hereafter referred to as anomalous and non-anomalous, respectively. To our knowledge, such a dataset does not exist, so in this section, we explain how the anomalous and non-anomalous observations were collected.
    
    Our H$\alpha$ anomaly dataset consists of anomalous and non-anomalous ground-based observations collected from the GONG Archive in FITS format and converted to JPEG format. For conversion, we used a Python wrapper \cite{mlecofi2024fits2jpeg} which utilizes the same algorithm used by the NSO to convert images \cite{fits2jpeg2021cotton}.

    Non-anomalous images were collected by repeatedly selecting 1,000 days (randomly) between 2011 and 2021, and choosing one observation per day. Solar activity may change significantly with the 11-year solar cycle. Using a 10-year period for selecting non-anomalous images allows us to mitigate that solar variation. The randomly collected images were then manually reviewed and any observation that exhibited anomalous patterns was moved to the anomalous set. Random selection was repeated until 2,000 H$\alpha$ images, with half of them being anomalous observations, were collected. This balanced dataset is used for the experiments, but in Sec.~\ref{subsec:impact_of_class_imbalance} we also examine the impact of class imbalance on performance, to address the extreme-imbalance nature of the task.

    The dataset was labeled manually by categorizing the following patterns on the solar disk as anomalous: objects obstructing the solar disk, shadow (clouds), an uncentered solar disk, oversaturation, and the absence of normal texture. The root-causes of some of these patterns are briefly described in Section \ref{sec:haobs}. An image without a normal texture could be related to observations taken through a relatively thick cloud with the solar disk still visible, but with any details on the disk washed out.

    Each image labeled as anomalous has at least one of the mentioned defects. A few examples of such cases are shown in Fig.~\ref{fig:anomaly_samples} where vegetation, objects, or clouds obstruct the solar disk, a filter is used on the lens as an ad-hoc measure of handling maintenance downtime, or the image quality has degraded during post-processing, resulting in an oversaturated appearance. The remaining observations, including those with mild shadowing, were classified as non-anomalous. Notably, image quality, whether sharp or blurry, was not a criterion for labeling. However, it is worth mentioning that many of the identified anomalies were often accompanied by some degree of blurriness. This dataset, namely the GONG H-Alpha Anomaly Dataset, is publicly available \cite{GONG_H_Alpha_Anomaly_Dataset_2025}.
    
    Let us briefly address some potential sources of bias in our data-collection process, and our mitigation strategies. First, every image returned by our random-selection algorithm was labeled either as anomalous or non-anomalous. This is important because otherwise, i.e., when some images are excluded from being labeled, the final dataset would not have been representative of the true population. That said, after we reached the desired number of non-anomalous observations, while still searching for anomalous observations, we had to ignore non-anomalous observations. To mitigate the potential impact of this exercise, we tried our best to examine whether similar patterns have already been represented by the already-collected images. If not, we replaced the new image with those that already have a good representation. Another source of potential bias lies in the fact that the collected set of anomalous observations may not be as comprehensive as it ideally should be. This is due to the scarcity of those images, and it may have resulted in some extremely infrequent anomalous patterns never being selected. Note that there is no information on the distribution of different types of anomalous images of the GONG network. We acknowledge this possibility, but we believe our qualitative analysis of the false alarms and misses, in Sec.~\ref{subsec:qualitative_analysis}, addresses this concern.

\section{Image Anomaly Detection Efforts}

    In this section, we will review some of the most commonly used anomaly detection methods and introduce at least one example of each that will serve as our baseline of comparison. From an algorithmic perspective, methods can be classified as shallow (e.g., One-Class SVM \cite{scholkopf_2001_OCSVM}) or deep (e.g., Autoencoders \cite{rumelhart_learning_1986}, GAN-based approaches \cite{goodfellow_generative_2014}). From a data availability perspective, they are often grouped into unsupervised, semi-supervised, and supervised approaches. Unsupervised anomaly detection is typically applied when no labeled anomalies are available. Many approaches in this setting fall under one-class classification, where the model is trained only on non-anomalous data. Semi-supervised anomaly detection appears in two forms: (1) training with a majority of normal instances and a small set of labeled anomalies, and (2) \textit{Learning from Positive and Unlabeled Examples} (LPUE), where only normal data are labeled and the remaining data are unlabeled. Supervised anomaly detection treats the problem as a standard classification task in which both normal and anomalous classes are fully represented in the training set. This scenario is uncommon in practice, as anomalous patterns are often rare \cite{ruff_2021_unifying}. 
    
    \subsection{Classic methods}
    The most popular classic anomaly detection methods include One-Class Support Vector Machine (OCSVM) \cite{scholkopf_2001_OCSVM} for unsupervised and Support Vector Machine (SVM) \cite{cortes_support-vector_1995} for supervised anomaly detection. 
    
    OCSVM maps the input space to a high-dimensional feature space using the kernel trick and determines the hyperplane that best encloses the normal instances while maximizing their separation from the origin. Eq. \ref{eq:ocsvm_objective} shows the objective function of OCSVM for a training set of $n$ non-anomalous instances, where $\Phi$ is the feature map induced by the kernel, $\mathbf{w}$ is the weight vector of the separating hyperplane, $\rho$ is the offset of the hyperplane from the origin, $\nu \in (0,1]$ represents the upper bound of fraction of outliers and lower bound of support vectors, and $\xi_i \ge 0$ are slack variables allowing margin violations for sample $i$.
    By using minimum volume set estimation in this new feature space and minimizing $\rho$, OCSVM ensures that at test time, anomalous instances will fall on the other side of the hyperplane.
    \begin{equation}
        \begin{aligned}
            &\min_{\mathbf{w}, \, \xi, \, \rho}
            && \frac{1}{2} \|\mathbf{w}\|^2 + \frac{1}{\nu n} \sum_{i=1}^n \xi_i - \rho \\
            &\text{subject to}
            && (\mathbf{w} \cdot \Phi(\mathbf{x}_i)) \ge \rho - \xi_i,  \xi_i \ge 0,  \forall i
        \end{aligned}
    \label{eq:ocsvm_objective}
    \end{equation}

    Another algorithm that we include in our investigation is SVM. Given a training set of $n$ anomalous and non-anomalous samples, we solve the objective function in Eq. \ref{eq:svm_objective} such that the decision function $f(\mathbf{x}) = \operatorname{sign} \left( (\mathbf{w} \cdot \mathbf{x}) + b \right)$ is correct for the largest possible subset of training instances. In the objective function, $\mathbf{w}$ is the weight vector of the separating hyperplane, $b$ is the bias term shifting the hyperplane, $\xi_i \ge 0$ are slack variables, and $C > 0$ is the regularization parameter controlling the trade-off between maximizing the margin and penalizing margin violation.
    \begin{equation}
        \begin{aligned}
            &\min_{\mathbf{w},\,b,\,\xi} \quad \frac{1}{2} \lVert \mathbf{w} \rVert^2 + C \sum_{i=1}^n \xi_i \\
            &\text{subject to} \quad y_i \big(\mathbf{w}^\top \mathbf{x}_i + b\big) \ge 1 - \xi_i, \xi_i \ge 0, \forall i
        \end{aligned}
    \label{eq:svm_objective}
    \end{equation}
    
    When using kernel functions such as the radial basis function (RBF), polynomial, or sigmoid, an additional hyperparameter $\gamma > 0$ controls the influence of each training sample. For the RBF kernel, for instance, $\gamma$ determines the width of the Gaussian such that large values lead to more localized decision regions, which can cause overfitting, while small values yield smoother and broader boundaries. In practice, setting  $\gamma = \frac{1}{n_{\text{features}}}$ referred to as ``auto" or $\gamma = \frac{1}{n_{\text{features}} \cdot \mathrm{Var}(X)}$ referred to as ``scale", to scale $\gamma$ based on data variance, are popular.
    
    \subsection{Deep Learning Methods}
    
    As in other problems, Deep Neural Networks (DNNs) have quickly become the state-of-the-art methods for anomaly detection. As discussed, anomaly detection methods often rely on only non-anomalous or very few anomalous samples, due to the scarcity of anomalous instances. To this end, unsupervised and semi-supervised DNN models have shown great strength in this task as their deep hierarchical architecture enables learning the underlying distribution of images. Generative Adversarial Networks (GANs) \cite{goodfellow_generative_2014} and Autoencoders \cite{rumelhart_learning_1986} are the major unsupervised DNN-based methods used for this purpose.
    
    GANs are popular unsupervised models that implicitly approximate the training data distribution by learning to produce samples indistinguishable from real ones by training two networks: a generator \textit{G} and a discriminator \textit{D}. The generator \textit{G} has to generate adversarial samples from the training data that the discriminator mistakenly classifies as being from the same distribution as the training data, hence learning the training data distribution. \textit{G} consists of an encoder and a decoder, reconstructing the image by the decoder from its representation in the latent space, generated by the encoder. When used for anomaly detection, GANs are typically trained only on normal images. However, since they do not provide anomaly scores directly, at test time, the anomaly score is calculated from reconstruction error and/or feature space discrepancies measured using intermediate layers of \textit{D} \cite{ruff_2021_unifying}.

    GANomaly is a GAN-based model proposed for anomaly detection \cite{akcay_ganomaly_2018}. The proposed architecture consists of a generator and a discriminator comprising three sub-networks in the generator: an encoder $G_E$, a decoder $G_D$, and a second encoder $E$. $G_E$ encodes an input image $X$ to the latent space $Z$. $G_D$ decodes $Z$ to reconstruct the image $X'$. $E$ encodes $X'$ back to the latent space $Z'$. The objective function of this network shown in Eq. \ref{eq:ganomaly_objective} consists of three loss functions, $\mathcal{L}_\text{adv}$ optimizing the discriminator to mistake real and fake images, $\mathcal{L}_\text{con}$ optimizing the generator to generate realistic images, and $\mathcal{L}_\text{enc}$ to reduce the difference of latent representation of $X$ and $X'$. Since this model is only trained on non-anomalous images, $\mathcal{L}_\text{adv}$ is expected to be high enough for anomalous images, resulting in the discriminator easily identifying them as anomalous.
    
    \vspace{-0.2cm}
    \begin{equation}
        \mathcal{L} = w_\text{adv}\mathcal{L}_\text{adv} + w_\text{con}\mathcal{L}_\text{con} + w_\text{enc}\mathcal{L}_\text{enc}
        \label{eq:ganomaly_objective}
    \end{equation}

    The works mentioned above have not yet been applied to the problem of finding anomalous H$\alpha$ observations, and their effect has not been studied. In this proof of concept, we aim to analyze and compare the effectiveness of unsupervised and supervised deep and shallow anomaly detection methods with our statistical approach, H-Alpha Anomalyzer.
    
    \subsection{Anomaly Detection on H$\alpha$ Observations}\label{subsec:anomaly_detection_on_halpha}
        As stated in Sec.~\ref{subsec:lack_of_dedicated_quality_control}, a quality-controlled pipeline for H$\alpha$ observation has not been developed; Nevertheless, the necessity for such a system is clear, as there have been scattered attempts to create one. An example of such efforts is seen in \cite{veronig_2000_automatic}, where it is emphasized that image pre-processing is needed before continuing with attempting to segment solar events from H$\alpha$ images captured by the Kanzelh{\"o}he Solar Observatory (KSO). Their proposed method is based on comparing the distribution of pixel intensities with the normal distribution; deviations from the standard distribution indicate low-quality images that should not be further processed and segmented. Similarly, a flare detection pipeline implemented at KSO first removes the anomalous images that show prominent defects \cite{potzi_real-time_2015}. This pipeline was further improved and updated later in \cite{potzi_event-based_2018}, where the outcome of pre-processing the observations is described as observations being categorized as good, fair, and bad. Bad observations (equivalent to the anomalous class in this work) are moved to a temporary archive, whereas fair images are saved for visual inspection.
        To our knowledge, none of the mentioned algorithms have been made publicly available, so the community can not leverage those solutions; therefore, we have not included them in our comparison.

        More recently, a new system for quality assessment at KSO has been utilized, which leverages a GAN architecture to find observations with low quality \cite{jarolim_image-quality_2020}. As mentioned above, the anomalous observations have already been removed, and this system only distinguishes between the fair and good classes. Although this code is publicly available, the solved problem differs from the issue at hand, so we have not included it in our comparative study.
        Similarly, other solar event detection pipelines have focused on related preprocessing steps, such as applying corrections \cite{riegler_filament_2013} or mentioning benefiting from preprocessing the data to remove images with high pixel intensity around filaments \cite{zhu_solar_2019}.

\section{Method}
\begin{figure}[t]
        \centering
        \includegraphics[width=0.85\linewidth]{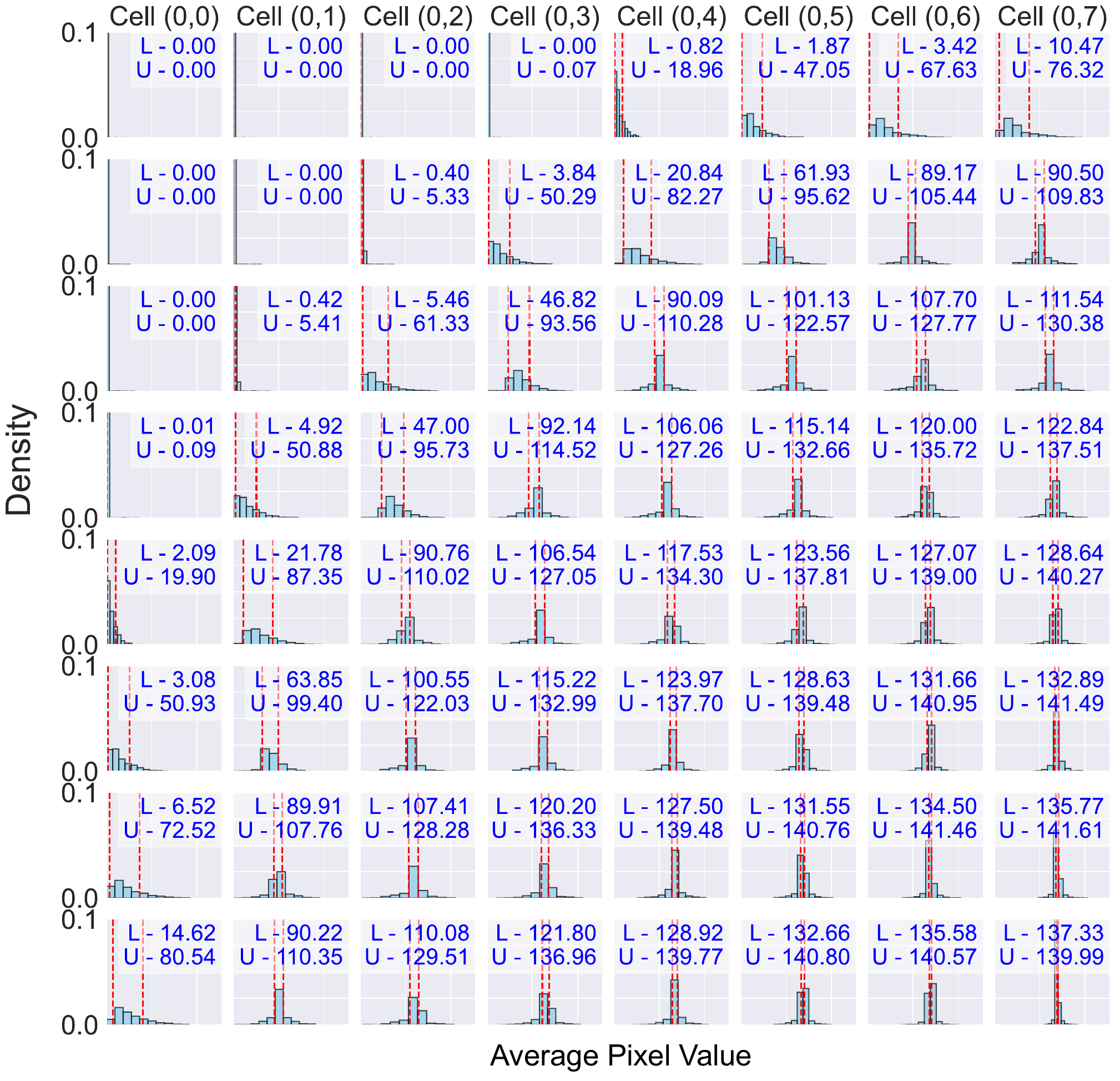}
        \caption{\footnotesize Density of average pixel value per cell in the training set shown for the first 8 cells (row-wise and column-wise). The two red dashed vertical lines mark the optimal upper range $U_{ij}$ and lower range $L_{ij}$ for that cell. The graph is best visible in digital format.}
        \label{fig:cell_pixel_avg_plot}
        \vspace{-0.4cm}
    \end{figure}

    Consider a training set $D_{\text{train}} = \{(x_k, y_k)\}_{k=1}^m$ consisting of anomalous and non-anomalous images, where $x$ denotes an image, and $y$ denotes its binary label; anomalous or non-anomalous. For a given image $x \in D_{\text{train}}$, we partition it into $n \times n$ cells indexed by $(i,j)$ where $i, j \in \{0, ..., n-1\}$, and define the statistic $I_{ij}$ as the average pixel-intensity of the $(i,j)$-th cell. Our proposed algorithm finds the distribution of $I$ per cell for anomalous and non-anomalous samples in $D$. Even though this distribution can be helpful, since we are averaging the pixel intensities in a cell, pixels with higher values (brighter) and lower values (darker) will all average to the range of gray. This will result in an overlap between the two distributions, which can confuse the model. In other words, we need to find a range of pixel intensities per cell that is considered normal and has the least overlap with the abnormal range. We define the upper and lower bounds of this range as $U_{ij}$ and $L_{ij}$, and search for a combination of $U_{ij}$ and $L_{ij}$ starting from the 20th percentile and 80th percentile, respectively.
    We measure how much the average pixel intensity $I_{ij}$ deviates from each candidate $U_{ij}$ and $L_{ij}$ by defining a new statistic namely, $S$ defined as:
    Our proposed algorithm's objective is to find an optimal range for $I_{ij}$ representing non-anomalous cells. Ideally, all $I_{ij}$ of anomalous cells are expected to fall outside this range. To achieve this, we consider an upper and a lower bound for each cell and introduce the $S$ statistic as $S_{ijk} = |U_{ij} - I_{ijk}| + |L_{ij} - I_{ijk}|$, where $U_{ij}$ and $L_{ij}$ are candidate percentiles for the upper and lower bounds of the normal range, for cell $(i,j)$ of image $x_k$. To find the optimal $U_{ij}$ and $L_{ij}$, we use the one-way ANOVA F-test \cite{welch_comparison_1951} on the $S$ statistic to determine a range for normal cell pixel intensity that has the least overlap with the range of anomalous pixel intensity. Whichever combination of $U_{ij}$ and $L_{ij}$ that yields the maximum $F$ statistic is chosen as the best range, indicating maximum separability between the normal and anomalous distributions. Fig.~\ref{fig:cell_pixel_avg_plot} shows the optimal upper and lower ranges for the collected training set. Once an optimal pair of $U_{ij}$ and $L_{ij}$ is found for each cell, we can use this range to find anomalous instances in the test set.

    Given a test set $D_{\text{test}} = \{(x'_k, y'_k)\}_{k=1}^P$, we can use the optimal range to identify anomalous cells in each image $x'$. For a given image $ x' \in D_{\text{test}}$, we partition it into an $n \times n$ grid and compute $S_{ij}$ for each cell, using the optimal $U_{ij}$ and $L_{ij}$. As mentioned, $S_{ij}$ measures deviation from the normal pixel intensity; hence, higher values of $S_{ij}$ correspond to a higher probability of a cell being anomalous. 
    Lastly, to convert $S$ into likelihood, we run the Sigmoid function, $p_{ij} = \frac{1}{1 + e^{-s_{ij}}}$. 
    A threshold ($\theta$) on the likelihood may be used to flag cells as anomalous. Similarly, a threshold on the minimum number of anomalous cells may be used to flag an observation as anomalous. In Sec.~\ref{sec:experiments_and_results} we find an optimal set of these two thresholds as well.

\section{Experiments and Results}\label{sec:experiments_and_results}
    In this section, we compare the performance of H-Alpha Anomalyzer with the introduced models and present a quantitative and qualitative analysis of our findings. \footnote{The source code is available at \\ \href{https://bitbucket.org/dataresearchlab/anomalyzer_icdmw25/}{https://bitbucket.org/dataresearchlab/anomalyzer\_icdmw25/}}

    \subsection{Model Training}
        All models in Experiments 1 and 2 were trained and tested on a dataset of randomly curated GONG $H\alpha$ observations between 2011 and 2016 containing 1,000 anomalous and 1,000 non-anomalous observations (for details, refer to Sec.~\ref{sec:dataset}). The training set differs based on the method, as H-Alpha Anomalyzer and SVM take anomalous and non-anomalous images as input, while the rest are unsupervised. The test set remains the same for all models. All hyperparameters (if any) were tuned for each model on the validation set. In the following text, we provide the details of our training configurations.
        
        For SVM and OCSVM, the images were fed to the model as $n$ flattened $32\times 32$ vectors, where $n$ is the number of images in $D_\text{train}$, and each element in the vector is the mean pixel intensity of that cell. The pixel intensities were normalized before passing to the model to be between 0 and 1.    
    
        \textbf{Ganomaly.} The Ganomaly model was trained using the Anomalib \cite{akcay2022anomalib} package's implementation. After applying random augmentation, we trained the model for 80 epochs on the train set of 700 non-anomalous images. The model was trained using these hyperparameters following GANomaly's recommendation \cite{akcay_ganomaly_2018}: $\text{learning rate}=0.0002,\text{batch size}=16,\beta_1=0.5,\beta_2=0.999$, and the following weights were used for the adversarial loss: $w_{adv}=1,w_{con}=50,w_{enc}=1$.

        \textbf{SVM.} SVM was trained on both anomalous and non-anomalous images as it's a supervised method. The optimal hyperparameters were determined via a grid search with $C \in \{10^{-2}, 10^{-1}, 10^{1}, 10^{2}\}$ , $\gamma \in \{\text{scale}, \text{auto}, 2^{-12}, 2^{-11}, ..., 2^{3}\}$ and kernel type chosen from $\{ \text{RBF}, \text{linear}, \text{poly}\}$.
        The optimal hyperparameters on the dataset for both Experiment 1 and Experiment 2 were found to be: $C=1,\gamma=0.25,\text{kernel}=\text{RBF}$.
        
        \textbf{OCSVM.} OCSVM was trained only on non-anomalous images. Similar to SVM, the optimal hyperparameters were determined via a grid search with $\nu \in \{0.01, 0.05, 0.1, 0.2, 0.5\}$, $\gamma \in \{\text{scale}, \text{auto}, 2^{-12}, 2^{-11}, ..., 2^{3}\}$ and kernel type chosen from $\{ \text{RBF}, \text{linear}, \text{poly}\}$.
        The hyperparameters on the dataset for experiment 1 were found to be $\nu = 0.5$, $\gamma =2^{-7}$, $\text{kernel}=\text{RBF}$, and for experiment 2 were found to be $\nu=0.2$, $\gamma = 2^{-10}$, $\text{kernel} = \text{RBF}$.
                
        \textbf{H-Alpha Anomalyzer.} Both anomalous and non-anomalous images were passed to the model to compute the range of pixel intensity of non-anomalous images. The following parameters were found to be optimal and used for the experiment: $\text{grid size}=16$, $\theta=0.7$, and $\text{minimum-corrupt-cells}=4$. 

    \vspace{-0.2cm}
    \subsection{Experiment 1: Performance Under Identical Conditions}
        We compare all listed models under identical conditions using the balanced dataset of 1,000 anomalous and 1,000 non-anomalous images. This dataset is split into training, validation, and test sets, with ratios of 0.7, 0.1, and 0.2, respectively. Each set has a 1:1 corrupt-to-normal ratio. It's important to note that this split remains the same for all methods, so all methods are tested against the same test set. The hyperparameters of SVM and OCSVM were optimized using grid search on the validation set to give them an advantage.
        
        The results of this experiment are presented in Table \ref{tab:test_result_balanced}. H-Alpha Anomalyzer and SVM have similar performance in terms of F1 score. GANomaly's performance is close to SVM but has a significantly higher false-negative rate (FNR). Even though the performance of OCSVM is not comparable to the other models, it is achieving the highest true-positive rate (TPR) by correctly classifying all anomalous samples, which, of course, comes at the cost of having a high false alarm (false-positive rate; FPR). 

        \begin{table}[]
            \centering
            \begin{tabular}{|l|l|l|l|l|l|l|}
            \hline
            Method           & TP  & TN  & FP  & FN & F1-Score \\ \hline
            GANomaly         & 183 & 197 & 3   &  17&  0.95    \\ \hline
            OCSVM            & 200 & 102 & 98  & 0  & 0.80     \\ \hline
            SVM              & 194 & 197 & 3   & 6  & 0.98     \\ \hline
            H-Alpha Anomalyzer & 192 & 196 & 4   & 8  & 0.97      \\ \hline
            \end{tabular}
            \caption{\footnotesize Performance comparison on balanced dataset}
            \label{tab:test_result_balanced}
            \vspace{-0.4cm}
        \end{table}

    \vspace{-0.2cm}
    \subsection{Experiment 2: Impact of Class Imbalance}\label{subsec:impact_of_class_imbalance}
        The anomaly detection task is, by definition, a class-imbalance problem. So, it is critical to validate the experiments with this assumption. However, the actual rate at which an observation becomes corrupt is unknown. This is because of the absence of an algorithm that can process a sufficiently large sequence of observations and estimate this rate. That said, the impact of class imbalance is minimal if the anomaly detection algorithm is very successful in identifying good observations, meaning, if it achieves a very high true-negative rate (TNR) (i.e., good observations that are correctly identified). To test this, we increase the corrupt-to-normal ratio and monitor the performance.
        
\begin{figure}
            \centering
            \includegraphics[width=1\linewidth]{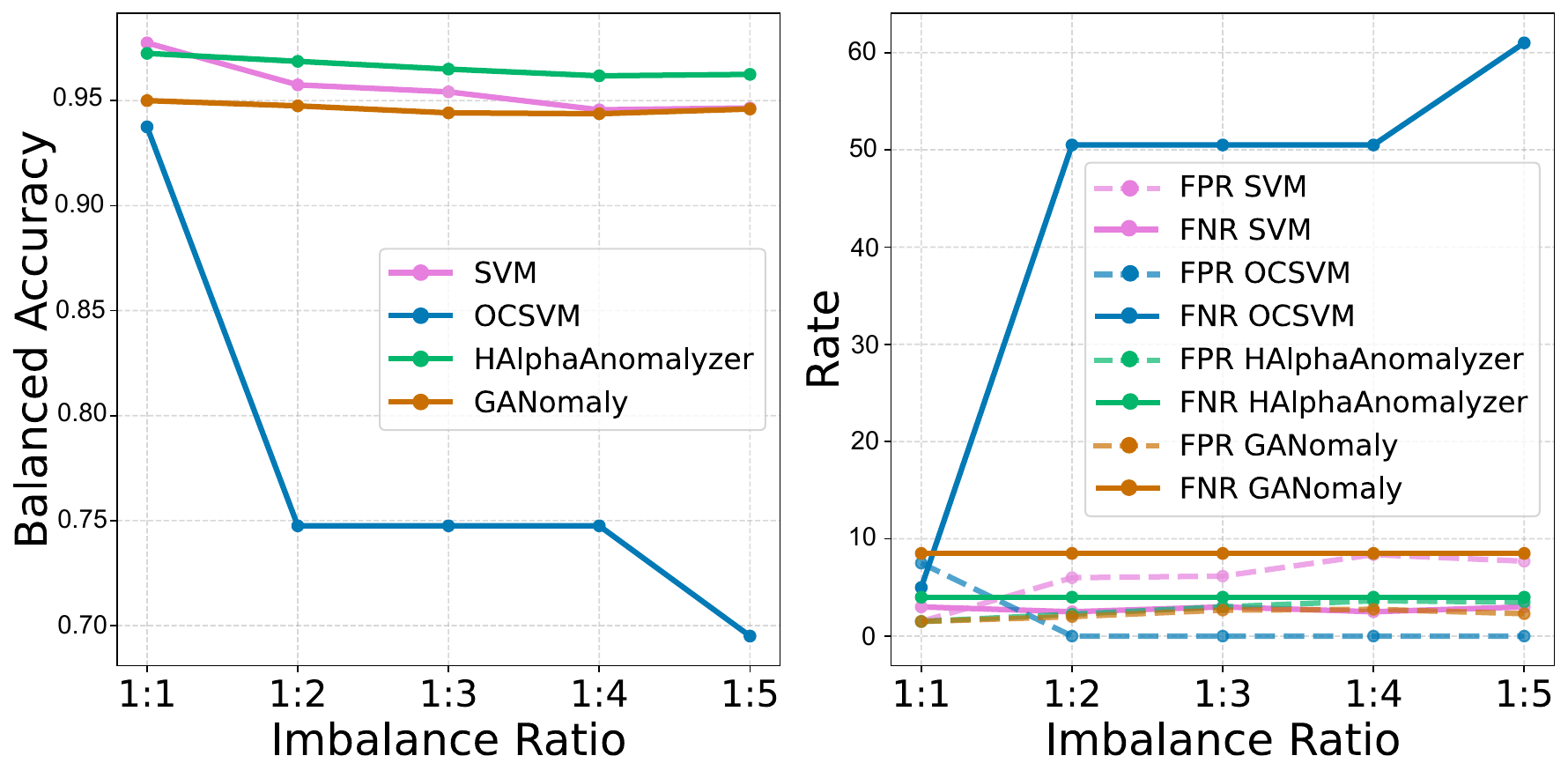}
            \caption{\footnotesize
            Effect of increasing the imbalance ratio on anomaly-detection models. The plot shows balanced accuracy, $\text{Balanced Accuracy} = 0.5(\text{Sensitivity} + \text{Specificity})$ where $\text{Sensitivity} = \frac{\text{TP}}{\text{TP} + \text{FN}}$ and 
            $\text{Specificity} = \frac{\text{TN}}{\text{TN} + \text{FP}}$.}
            \label{fig:accuracy_imbalanced_experiment}
            \vspace{-0.4cm}
        \end{figure}

        For this purpose, we gradually increase the class-imbalance ratio from 1:1 to 1:5 corrupt-to-normal images. To do this, we had to add more non-anomalous samples to our dataset. These new images were sampled in the same fashion as discussed in Sec.~\ref{sec:dataset} but from the years between 2017 and 2025, to avoid overlap with the previously sampled images. The results are shown in Table~\ref{tab:test_result_imbalanced}, as well as in Fig.~\ref{fig:accuracy_imbalanced_experiment}. One could see a natural decrease in performance of all models when the dataset is imbalanced; however, the decrease is more subtle in the case of H-Alpha Anomalyzer and GANomaly. SVM's FPR increased from 1.5\% to 8\%, while H-Alpha Anomalyzer's FPR increased from 2\% to 3.5\%. Looking at the H-Alpha Anomalyzer's anomaly scores, we could visually verify that in 71\% of the newly misclassified instances, the anomaly is detected in the ring around the solar disk (see Fig.~\ref{fig:anomaly_scores}-d). Even though not verifiable by anomaly scores, this pattern is also seen in all misclassified samples by SVM (77 false positives). As this ring is an artifact of on-site post-processing of H$\alpha$ images, the 3.5\% FPR for our method could be reduced even more by simply removing this type of anomaly from our analysis. Such a pattern is easy to identify because, in otherwise-normal observations, the ring always spans over a fixed set of grid cells. For OCSVM, however, performance drops by 28\% and FPR increases by 37.5\%. Similar to H-Alpha Anomalyzer, GANomaly shows little sensitivity to imbalance; however, as it does a better job in identifying the negative (non-anomalous) class, it has a higher FNR than H-Alpha Anomalyzer. Overall, H-Alpha Anomalyzer is the least impacted method by increasing the imbalance ratio and achieves the best balanced accuracy.

        \begin{table}[]
            \centering
            \begin{tabular}{|l|l|l|l|l|l|l|}
            \hline
            Method            & TP  & TN  & FP  & FN & F1-Score \\ \hline
            GANomaly          & 183 & 977 & 23  & 17 & 0.90      \\ \hline
            OCSVM             & 200 & 625 & 375 & 0  & 0.52      \\ \hline
            SVM               & 194 & 923 & 77  & 6  & 0.82      \\ \hline
            H-Alpha Anomalyzer& 192 & 965 & 35  & 8  & 0.90      \\ \hline
            \end{tabular}
            \caption{\footnotesize Performance comparison on imbalanced dataset with 1:5 corrupt-to-normal ratio.}
            \label{tab:test_result_imbalanced}
            \vspace{-0.5cm}
        \end{table}
        
    \subsection{Qualitative Analysis}\label{subsec:qualitative_analysis}
        In this section, we focus on both type I and type II errors of the algorithms, i.e., false-positive and false-negative rates (FPR and FNR), respectively. We investigate what anomalous patterns are not identified and what normal patterns are confused as anomalous.
        
        \textbf{SVM.} SVM yields the highest F1 score in experiment 1, and looking at the misclassified samples, we could not identify any specific pattern. The missclassified anomalous patterns by SVM are uniformly distributed among oversaturated, cloudy, and obstructed observations. These misclassifications are not systematic; for example, the misclassified obstructed observations make up 15\% of all obstructed observations in the dataset. Similarly, two out of three missclassified non-anomalous observations show subtle anomalies that could have contributed to this classification.
        
        \textbf{OCSVM.} Looking at Tables \ref{tab:test_result_balanced} and \ref{tab:test_result_imbalanced}, we can see that OCSVM favors the FNR (type II error) over the FPR (type I error). This leads to anomalous images always being correctly classified while misclassifying the non-anomalous class in more than 50\% of the cases, which means that to have a low FNR, OCSVM pushes the decision boundary into the hyper-space of the normal class. As expected, no specific pattern in non-anomalous images is misclassified as anomalous, including misclassifying a mix of crisp, mildly cloudy, and blurry images belonging to the non-anomalous class.

        \textbf{GANomaly.} GANomaly has the highest FNR among all models, 52\% of which are coming from missclassifying observations obstructed by nearby objects, obscuring the field of view (e.g., Fig.~\ref {fig:obstructed}). Moreover, GANomaly is the only model failing to capture the uncentered disk and texture of the observation. That said, this model consistently has the lowest FPR across all imbalance ratios in Fig.~\ref{fig:accuracy_imbalanced_experiment} in exactly three instances, two showing subtle cloudy patterns.
        
\begin{figure}
        \centering
        \begin{subfigure}{.84\linewidth}
                \centering
                \includegraphics[width=1.0\linewidth]{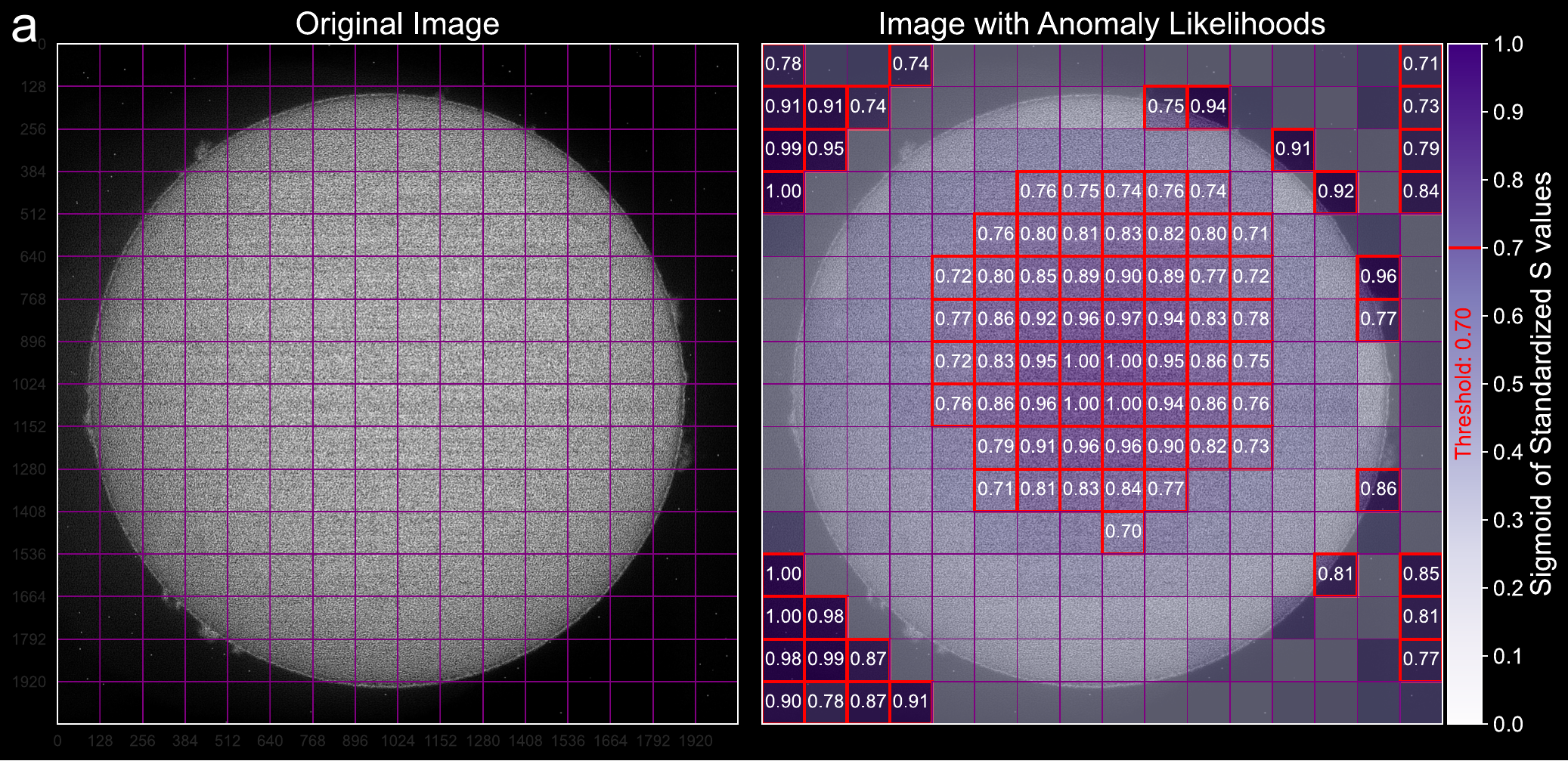}
        \end{subfigure}

        \begin{subfigure}{.84\linewidth}
                \centering
                \includegraphics[width=1.0\linewidth]{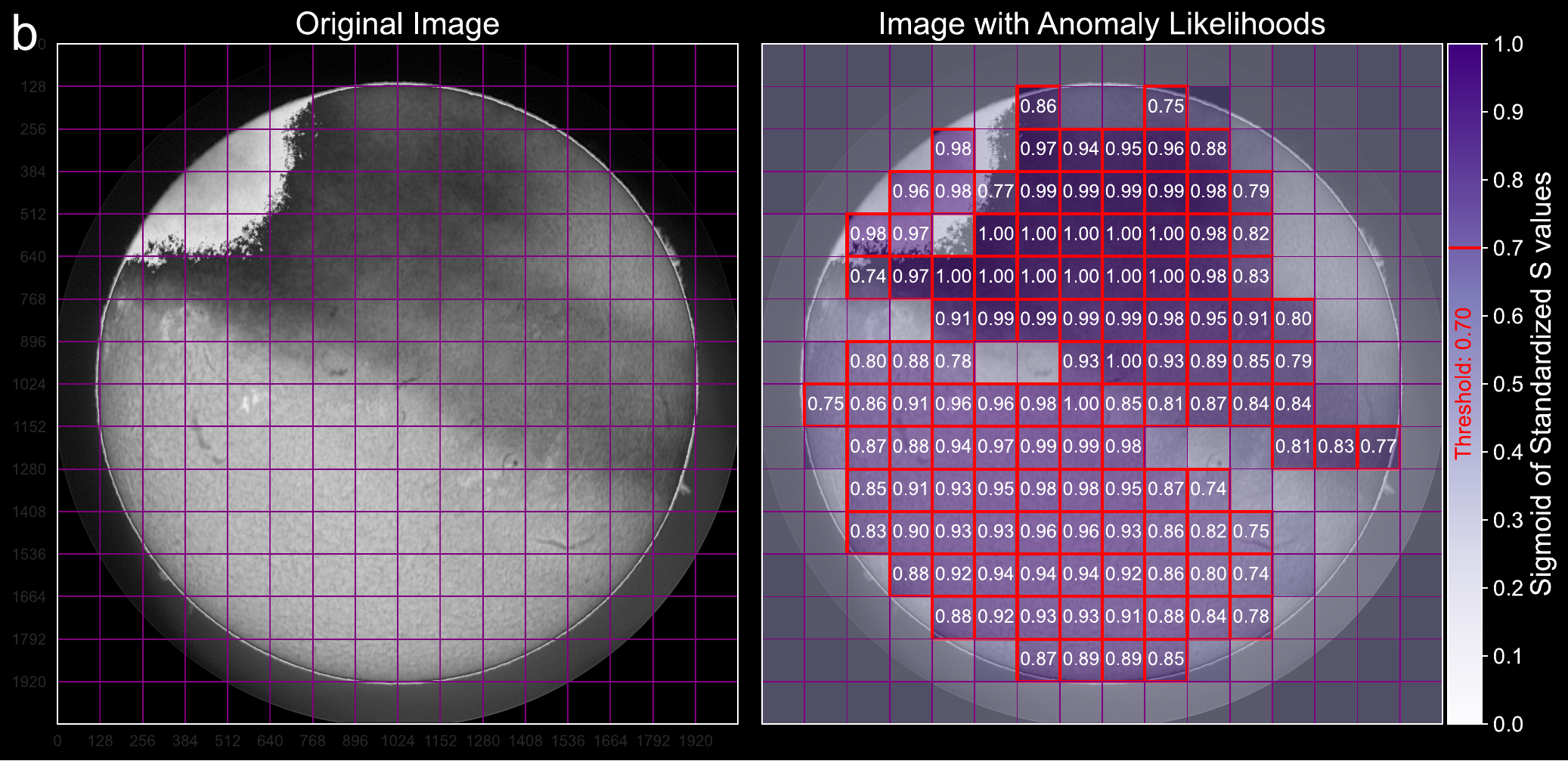}
        \end{subfigure}

        \begin{subfigure}{.84\linewidth}
                \centering
                \includegraphics[width=1.0\linewidth]{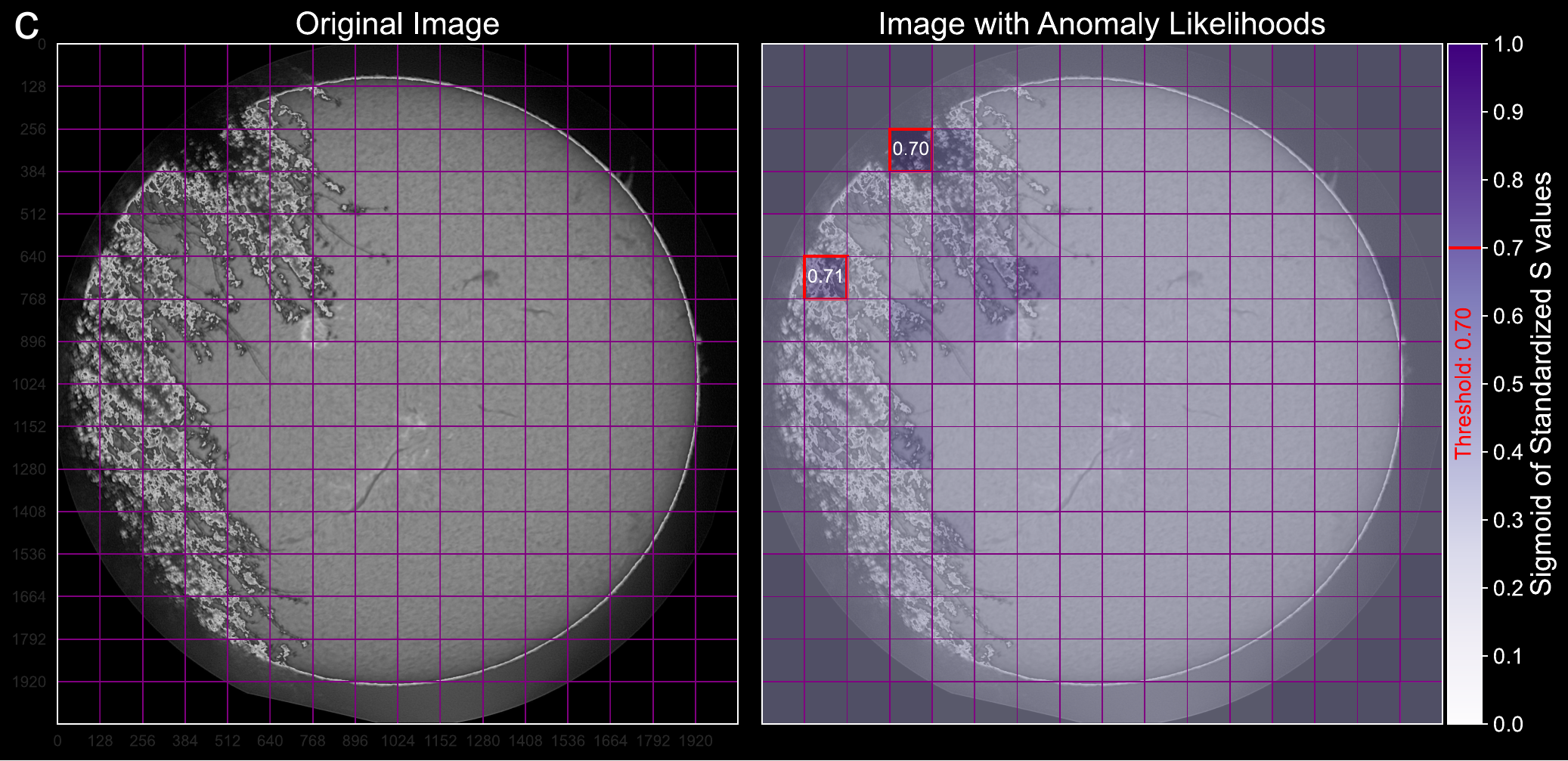}
        \end{subfigure}
        
        \begin{subfigure}{.84\linewidth}
                \centering
                \includegraphics[width=1.0\linewidth]{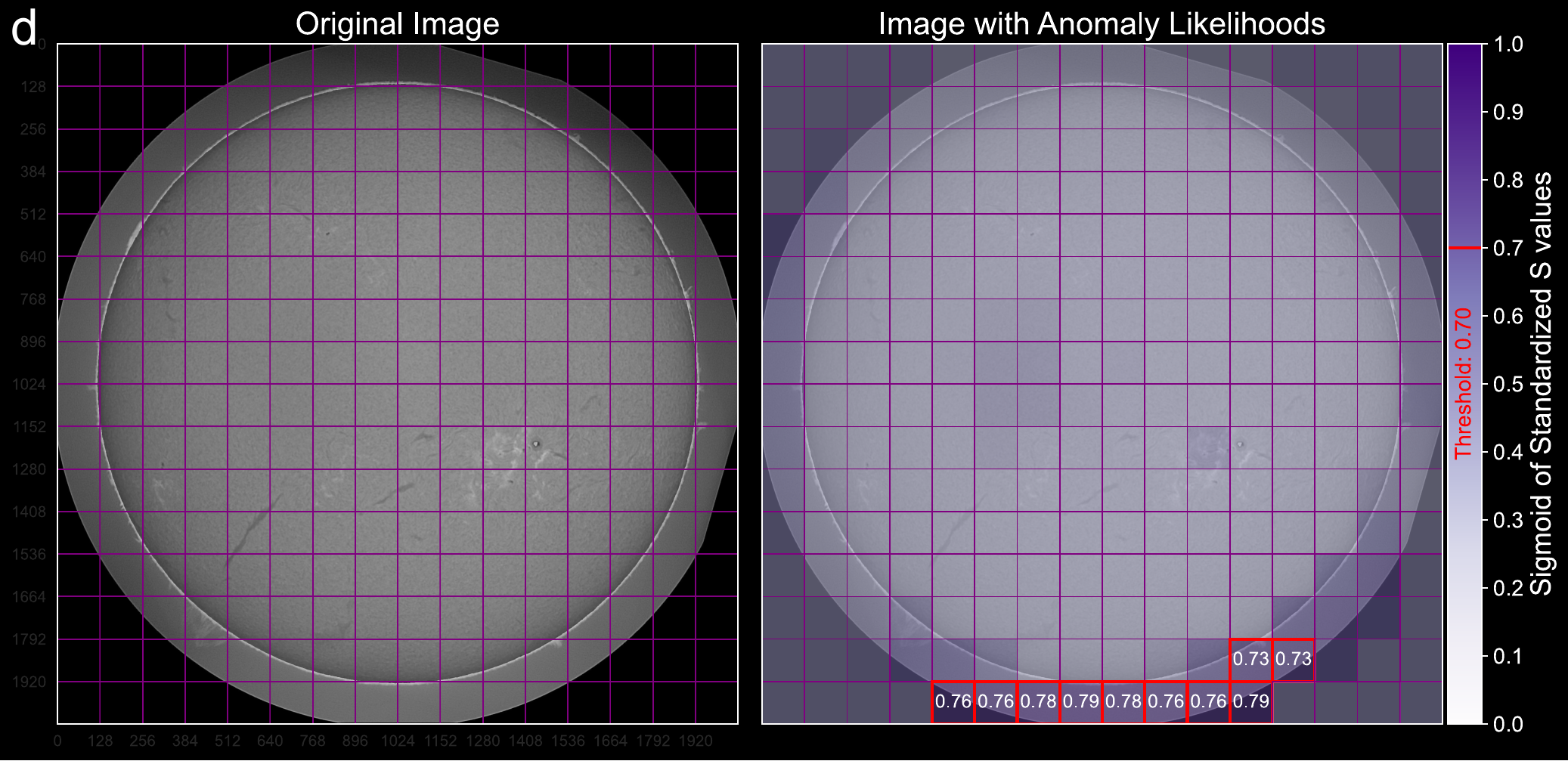}
        \end{subfigure}
        
        \caption{\footnotesize Examples of H-Alpha Anomalyzer outputs with anomaly scores. Timestamps of observations from (a) to (d) \mbox{are: 
        20110403142554Bh, 20150604093554Uh, 20220405000250Bh,} \mbox{20160124024354Uh}}
        \label{fig:anomaly_scores}
        \vspace{-0.5cm}
        \end{figure}

        \textbf{H-Alpha Anomalyzer.} H-Alpha Anomalyzer is comparable to SVM and GANomaly in terms of its F1 score; however, since this method provides explainability through giving the anomaly scores for each cell (see Fig.~\ref{fig:anomaly_scores}), we can dive deeper into its weaknesses and strengths. Figs. \ref{fig:anomaly_scores}-a, \ref{fig:anomaly_scores}-b show correctly classified samples, and \ref{fig:anomaly_scores}-c, \ref{fig:anomaly_scores}-d show samples misclassified by this method. 

        One interesting correctly classified observation is Fig.~\ref{fig:anomaly_scores}-b, where some of the oversaturated pixels resulting from post-processing of GONG images are being canceled out by the black pixels in the same cell in the background, which explains why cells containing the white pixels near the limb are classified as normal. What contributes to this image being classified as anomalous is the dark shadow inside the ring and a brighter region in the southwest of the disk. At first, this might look like a mistake; however, a closer look at normal GONG images reveals that this region is typically not this bright. This brightness could be an artifact of the post-processing.
                
        Looking at the misclassified instance \ref{fig:anomaly_scores}-c, it is quite obvious that this image is an anomaly; however, considering that H-Alpha Anomalyzer works with mean pixel intensities, it becomes clear where the misclassification stems from. The corrupt cells in this image exhibit a static-noise-like pattern, which results in the pixel intensities averaging at a gray level that is considered ``normal'' for those cells. Many cells in this observation should have been marked as anomalous, but only two were.

        A closer look at misclassified patterns identified by H-Alpha Anomalyzer reveals that out of the 200 anomalous images in the test set, about 3\% of cases with oversaturation, 8\% of cases with a cloudy pattern, and 31\% of obstructed cases were misclassified. SVM rates are as follows: 5\% of cases with oversaturation, 4\% of cases with a cloudy pattern, and 15\% of obstructed cases. In fact, 5 out of 8 instances misclassified by our method were misclassified by SVM too. We believe this is due to the effect of using mean of pixel intensities for both SVM and H-Alpha Anomalyzer as described above, and not related to the algorithm.
    
        In summary, the H-Alpha Anomalyzer maintains the balance between FPR and FNR while being minimally affected by increasing the imbalance ratio. GANomaly seems to have difficulty identifying anomalous patterns (specifically obstruction) the most out of all models, and hence is suitable for keeping FPR low, that is, to inevitably have a higher FNR.

\section{Discussions and Future Work}\label{sec:discussion}
    In this paper, we introduced a statistical method, namely H-Alpha Anomalyzer, for finding anomalous H$\alpha$ observations and providing explainability for each choice. This method was compared to classic machine learning anomaly detection techniques, i.e., SVM and One-Class SVM, and black-box DNNs, to see how they compare in terms of performance. Our findings show that H-Alpha Anomalyzer has performance comparable to the SoTA methods while having the advantage of explainability, and hence can be used for automating the data cleaning process in quality control pipelines.

\section*{Acknowledgment}
    This material is based upon work supported by the National Science Foundation under Grant No. 2209912, 2433781, and 2511630 directorate for Computer and Information Science and Engineering (CSE), and Office of Advanced Cyberinfrastructure (OAC). This work utilizes GONG data obtained by the NSO Integrated Synoptic Program, managed by the National Solar Observatory, which is operated by the Association of Universities for Research in Astronomy (AURA), Inc.

\bibliographystyle{splncs04}
\raggedright
\vspace{-0.3cm}
\bibliography{main}

\end{document}